\documentclass{article}
\usepackage{graphicx}
\usepackage{amsmath,amssymb} %
\usepackage{color}
\graphicspath{{./figs/}}
\usepackage{times}
\usepackage{xspace}
\usepackage{placeins}
\usepackage{epsfig}
\DeclareMathOperator*{\argmin}{arg\,min}
\usepackage{capt-of}
\usepackage{setspace}
\def\eg{{\em e.g.}\@\xspace}
\def\etal{{\em et al.}\@\xspace}
\def\etc{{\em etc.}\@\xspace}

\usepackage[width=122mm,left=12mm,paperwidth=146mm,height=193mm,top=12mm,paperheight=217mm]{geometry}
\begin{document}

\title{Faster RER-CNN: application to the detection of vehicles in aerial images}  

\author{Jean Ogier du Terrail\footnote{This work was supported by ANRT through the CIFRE sponsorship No
2015/1008 and by SAFRAN Electronics and Defense.}\\
Safran Electronics and Defense\\\\
{\tt\small jean.ogier-du-terrail@safrangroup.com}
\and
Frederic Jurie\\
Normandie Univ, UNICAEN, ENSICAEN, CNRS\\
{\tt\small frederic.jurie@unicaen.fr}
}

\title{Faster RER-CNN: application to the detection of vehicles in aerial images}

\date{September 21, 2018}
\maketitle

\begin{abstract}
Detecting small vehicles in aerial images is a difficult job that can be challenging even for humans. Rotating objects, low resolution, small inter-class variability and very large images comprising complicated backgrounds render the work of photo-interpreters tedious and wearisome. Unfortunately even the best classical detection pipelines like \cite{fasterrcnn} cannot be used off-the-shelf with good results because they were built to process object centric images from day-to-day life with multi-scale vertical objects.
In this work we build on the  Faster R-CNN approach to turn it into a {detection} framework that deals appropriately with the rotation equivariance inherent to any aerial image task. This new pipeline (Faster Rotation Equivariant Regions CNN) gives, without any bells and whistles, state-of-the-art results on one of the most challenging aerial imagery datasets: VeDAI \cite{vedai} and give good results w.r.t. the baseline Faster R-CNN on two others: Munich \cite{munich} and GoogleEarth \cite{google_earth}.
\end{abstract}

\section{Introduction}

In just a few years the detection capabilities of modern methods have rocketed. From the Viola-Jones detector to the brand new installment of the R-CNN family \cite{maskrcnn} the computer vision community {in detection} has progressively let go of old paradigms (sliding windows, feature extraction$+$training, traditional NMS) to embrace shinier new ones (Fully convolutional methods, end-to-end, soft/learnable NMS), which has led to drastic improvements in speed, performances and flexibility in modern detection pipelines.

However it seems that one of them, the use of PASCAL VOC vertical bounding boxes for every single detection problem, is still resisting the course of time even in situations in which it makes little sense. 

There are two main reasons why it is the case; First of all, natural objects on images can have arbitrary shapes  therefore we either find a simple approximation that kind of work in every case (the vertical bounding boxes) or we move to semantic segmentation (or even do both like in \cite{maskrcnn}). Second, the need for a standard to be able to compare different methods on an equal footing with each other. Thanks to PASCAL VOC IoU criteria (see \cite{pascalvoc}) we can for instance compare the recall of very different object proposals.

\begin{figure}
\centering
  \includegraphics[width=0.5\linewidth]{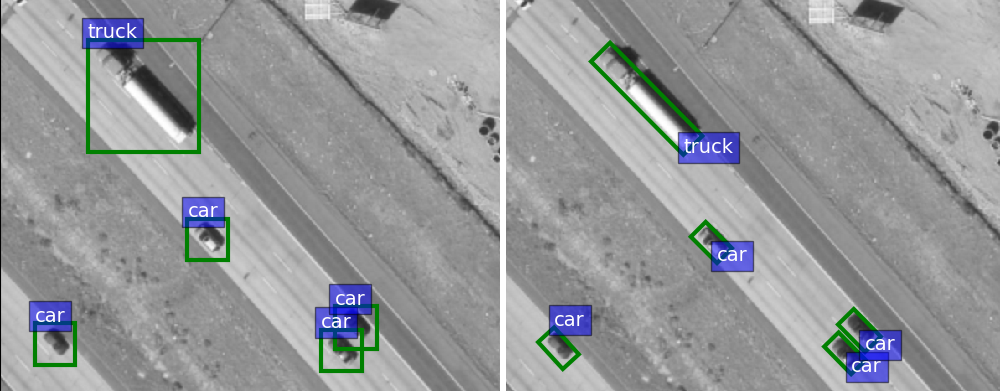}
 \caption{The two frameworks side by side: straight and oriented annotations}
\label{fig:test}
\end{figure}

We argue that in the specific domain of vehicle detection in aerial imagery these two reasons are lapsed. First, all the vehicles to detect seen from the sky are perfect rectangles or at least fairly close. That is why doing semantic segmentation like in \cite{audebert2} is arguably a bit 'overkill'. Moreover these rectangles can have any orientation (see Figure~\ref{fig:test}). 

It is therefore completely natural to propose a new paradigm: oriented rectangle-shaped bounding boxes for {annotations} (which already exist or have been provided in all aerial imagery datasets), for {region proposals} and for {evaluation} also making it a new unified coherent framework. 

This very simple idea has been, to the best of our knowledge, mostly ignored in aerial imagery until the very recent introduction of DOTA\cite{Xia_2018_CVPR} although it has received some attention by concurrent work very recently in the text detection community (\cite{rot_cvpr2017,rot_cvpr2017_2,Busta_2017_ICCV,ma2018arbitrary}) specifically for the ICDAR challenges. This led to authors either {completely ignoring orientations} by using vertical bounding boxes \eg, \cite{Tang2017} and thus inventing their own evaluation metric in the process (which prevents any fair comparison), {studying the orientations completely separately from the detections} \eg, \cite{komo_ri}, or at best studying the two problems simultaneously but {performing detection first on one side and only then angle regression on another} like in \cite{Liu15:fast}, a mindset which is visible in the different metrics used to evaluate the final result. Yet we prove experimentally that performing both simultaneously even improves the classification accuracy.

This contribution is therefore three-folded: 
(i) We adapt the Faster R-CNN framework to simultaneously do detection and orientation inference.
(ii) We suggest a suitable metric to compare rotated and non rotated detection methods
(iii) We get state-of-the-art results on a well-known aerial imagery dataset using the previous framework and get competitive performance w.r.t. the baseline on two others. %
\section{Related work}

General detection on images is a broad subject. Countless work has been done on the topic. However without going back all the way to Viola and Jones~\cite{viola} we feel compelled to refer to the emblematic series of the R-CNN articles \cite{fastrcnn,rcnn,maskrcnn,fasterrcnn}, the recent extensions like \cite{rfcn} and also all modern single shot methods like \cite{ssd,yolo,yolo9000}.

All those articles have inspired to various extents this paper the closest one being Faster R-CNN \cite{fasterrcnn}, which we extend in this work.

Vehicle detection on aerial imagery was also heavily investigated. Previous works like \cite{ali2012real,eikvil2009classification,kembhavi2011vehicle}
are worth mentioning. More recent studies involve handcrafted descriptors \cite{gleason_vedai,sebastien2} and/or convolutional neural networks (CNNs) \cite{chen2013vehicle,cowc}.

Lately in 2017, several articles in the domain stood out. Among them one can cite the article by Tang \etal \cite{Tang2017}, which demonstrates the need to extend the capabilities of Faster R-CNN in order to tackle aerial imagery detection. They chose to focus on augmenting the resolution of the feature maps while doing hard example mining via a boosting strategy. Our approach is orthogonal to their work so they could be combined with the benefits of both frameworks.

Audebert \etal \cite{audebert2} has also shown how important could segmenting vehicles correctly before classifying them be. But one needs a huge amount of pixel level data to even start talking about semantic segmentation, data which is not always available specifically in this domain. But these results are in concert with ours: once you have the orientation mask of the vehicle, classification is considerably simplified.

Ogier du Terrail and Jurie~\cite{me} got good results on VeDAI using a LeNet-5 \cite{LeCun:1998hy} fully convolutional network with hard examples mining but agree on the interest of a cascade for this problem.

Our work is meant as a tribute to Liu and Mattius~\cite{Liu15:fast}, who pioneered our direction of studies in aerial imagery they were the first to observe that the classification of cars is made easier if one lowers the intra-class variability by learning an angle regressor first and only then a classifier on the now vertical cars.

We feel that although many great recent articles worked on rotation invariant classifiers (possible first step in making a detector equivariant) like~\cite{komo_ri,boltzmann_ri} this line of work has been insufficiently explored especially in the context of detection. We note that Marco~\etal\cite{komo_ri} and Joao and Vedaldi~\cite{warped} learn to estimate the angles of vehicles from aerial imagery without considering the detection problem.

Interestingly, this is in the domain of {\em text detection} that we can find the most relevant literature although it seems still very recent. Both \cite{rot_cvpr2017,rot_cvpr2017_2} use oriented proposals and interior point methods like us. For text recognition it was abundantly clear from the start that one needs a detection aligned with the text and therefore that vertical bounding boxes are insufficient. However oriented rectangles are also too coarse an approximation. Liu and Jin~\cite{rot_cvpr2017} have to use complicated quadrangles and Shi \etal~\cite{rot_cvpr2017_2} have to resort to segment detection but even then both methods fail to capture text with too much curvature. In aerial imagery we do not have this problem nor the requirement to use quadrangles/segments instead of simple rectangles. That makes our method faster and more accurate as our prior can be stronger.  

We also built on the two very recent works of \cite{Busta_2017_ICCV,ma2018arbitrary}. However, even if the structure of their detector bears resemblance to ours they made very different design choices:
\begin{itemize}
\item Both of them use a simple regression cost, which did not converge in our case. It forced us to put more thoughts on engineering a more stable cost (see \ref{reg}).
\item Ma \etal \cite{ma2018arbitrary} parametrization is impractical because they need to recompute all corners for each bounding box during the IoU calculations whereas our allow implementation tricks (see \ref{iou}).
\item Busta \etal \cite{Busta_2017_ICCV} do not mention the NMS step nor the IoU computation it is unclear how they did it.
\item Both the bilinear interpolation from \cite{Busta_2017_ICCV} and the RRoI from \cite{ma2018arbitrary} are more precise but slower than our coarse pooling implementation that check pixel membership using dot products (see \ref{rroi})
\item They have different goals than us: Ma  \etal \cite{ma2018arbitrary} is doing oriented text detection and Busta \etal \cite{Busta_2017_ICCV} is more focused on text recognition.
\end{itemize}

Outside of the text detection domain, we have found very few articles that approach the detection problem with rotated proposals except  the  work from Chen and Dong \cite{msra_faces} on faces that can detect only vertical bounding boxes even if, thanks to the spatial transformer network \cite{stn}, they can warp to a certain extent what is inside the vertical regions to ease classification, which is close in spirit to this work.

\section{Adaptation of Faster R-CNN to orientation inference}

As aforementioned, the Faster R-CNN detector has been designed for datasets such as MS-COCO for which bounding boxes are not oriented.  This section proposes an efficient way to extend it to the prediction of oriented detection boxes. We first provide some minimum information about Faster R-CNN, give a parametrization of the bounding boxes allowing them to rotate, explain how to compute bounding boxes intersection in this context and finally how to extend the Faster R-CNN to allow it to handle rotating bounding boxes.

\subsection{Faster R-CNN, a reminder}
Faster R-CNN \cite{fasterrcnn} is a modern cascaded detection framework whose simplicity and robustness have seduced a wide range of authors since its coming of age in 2015. This pipeline is made of two parts. The first one, so-called RPN (Region Proposal Network) is used to propose (vertical) regions. In order to do so the image under study is divided into a grid of NxN cells. Each cell is responsible for a certain number of reference bounding boxes called {\em anchors}. Each anchor receives an objectness score and is shifted and distorted to better match the ground truth (this step is called bounding box regression). After scoring, a certain number of these new bounding boxes is then passed on (after Non-Max Suppression) to a second network.  

This second network called Fast R-CNN uses the regions to zoom in on interesting parts of the image feature maps, by a warping pooling process known as Region Of Interest Pooling. After RoI pooling all the regions have become fixed size features that can be further processed by fully connected layers to do fine grained classification (and can also be shifted and distorted a second and final time by a second layer of bounding box regression).  

This work gives more flexibility rotation wise to this framework by extending several key components. The first step in building this enhanced Faster R-CNN is rethinking how to parametrize rotated boxes in order to have a coherent framework and be able to efficiently compute IoU between those boxes, which is essential in Faster R-CNN.
We do so in the next section.

\subsection{Introducing the foundations of Faster RER-CNN}{\label{foundations}}

\subsubsection{Parametrizing rotating bounding boxes}{\label{parametrizing}}
The well know parameterization used for a vertical bounding box is:
\begin{equation}
\mathcal{B}=(x_{min},y_{min}, x_{max}, y_{max}),
\end{equation}
which corresponds to the top left corner and the bottom right-hand corner of the box.
Obviously allowing the boxes to rotate add degrees of liberty and therefore new parameters.

We wanted to find the simplest parameterization and the closest to the former so that the transition from one system to the other is as simple as possible. We were also motivated by the need to unify the annotations that can be found in aerial imagery datasets in surprisingly complex forms : the annotations  provided by Joao and Vedaldi in \cite{warped} for the GoogleEarth dataset \cite{google_earth}, for instance, is a perfect example of how authors in aerial imagery frequently did not really consider the detection and angle regression to be related, the angle annotations of VeDAI can also be confusing at first with some vehicle having orientations in $[0,\pi]$ and others in $[-\pi,\pi]$.

We chose the following:
\begin{equation}
\mathcal{B}=(x_{A},y_{A}, x_{C}, y_{C}, \theta),
\end{equation}
where A is one of the corners of the rectangle (A is chosen to be on the back of the vehicle if it has a back so that AB is the largest side, but it can be any corner {as long as this choice remain coherent throughout the annotations}). ABCD are the corners of the rectangle in clockwise order. We will see that it is crucial to get good performances. $\theta$ is the angle between the horizontal and AB counted clockwise

This parametrization encompasses all use cases that can be found in aerial imagery.
Furthermore this system has the nice property of being exactly the same as VOC if and only if $\theta=0$. This is therefore different from \cite{rot_cvpr2017_2}, which uses $w$ and $h$ instead of $x_{C}$ and $y_{C}$ and very different from the very complicated ten points coordinates of \cite{rot_cvpr2017} that certainly allows higher flexibility for text but is useless for aerial imagery.

\subsubsection{Pascal VOC IoU with oriented rectangles}{\label{iou}}
The Intersection Over Union is not only a way of asserting the compatibility between a predicted bounding box and the ground truth but also a crucial step in almost all modern pipelines (assessing if several boxes can be merged during NMS for instance).
We wanted to design an algorithm capable of keeping this simplicity without being too computationally intensive.
To do that we first compute the width and height of the two considered rectangles. Simple maths gives:
\begin{equation}
\begin{aligned}
AB&=\cos(\theta)*\Delta_{x}+\sin(\theta)*\Delta_{y} \\
BC&=-\sin(\theta)*\Delta_{x}+\cos(\theta)*\Delta_{y}
\end{aligned}
\end{equation}

with: $\Delta_{x}=x_{C} - x_{A}$ and 
 $\Delta_{y}=y_{C} - y_{A}$, where AB and BC are always positive with this parameterization (no need to take absolute value).  

Therefore, we can have access to the area of the individual rectangles. For the intersection, we need some more complex calculations. As all the rectangles are labeled in clockwise order, we can use the Sutherland-Hogman algorithm \cite{sh} to find the corners of the intersection by consecutively/successively clipping the first rectangle with each edge of the second one.  

Once we have the intersection -- which is therefore also in clockwise order -- we need only apply the Green-Riemann theorem \cite{gr} to have a simple and efficient way of computing the area of the intersection. 

\begin{figure}[tb]
\begin{center}
\includegraphics[width=0.4\linewidth]{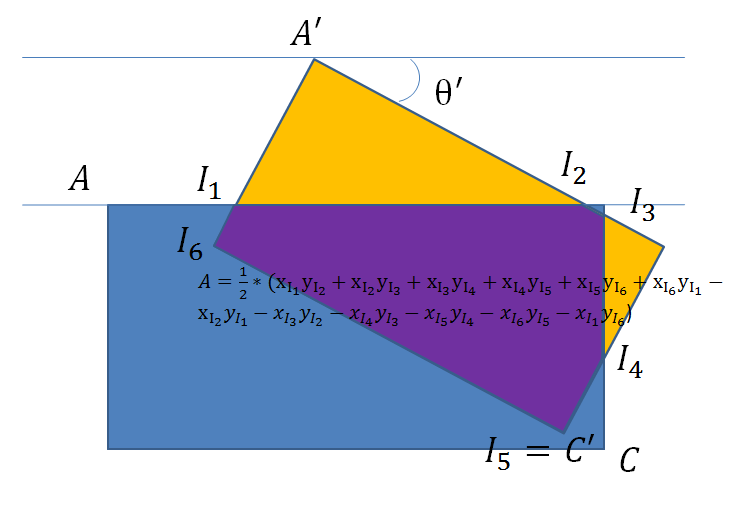}
\caption{Calculating the intersection of two oriented rectangles (note that in image coordinates all points are in anticlockwise order)}
\label{iou_rectangles}
\end{center}
\end{figure}

Thus we can calculate the IoU of two oriented regions in a non-painful way.
This is summed up in the Figure \ref{iou_rectangles}, which shows the GR formula's result from the corners coordinates $I_{1}$ to $I_{6}$.
Two ultra-efficient multithreaded Cython implementations of the IOU-calculation AND the new rotated NMS have been written.
The code needs to be multithreaded and very well thought-out otherwise the computation cost is too high, that is probably why the authors from \cite{rot_cvpr2017} favored Monte-Carlo simulations. Our algorithm has the advantage that it is exact.

Now the stage is set, we have what we need to move forward and are now ready to grapple with the actual implementation.

\subsection{From Faster R-CNN to Faster RER-CNN}
We will list in the following sections what are extensions of Faster R-CNN required to carry it forward to an aerial imagery setting with rotated objects.
\subsubsection{The RPN}
The first part of the R-CNN cascade is the region proposal part. To adapt the Faster R-CNN to our framework we simply duplicate the anchors of Faster R-CNN a fixed number of times in all directions between $[0,\pi]$ so that they span the entire spectrum of orientations (in order not to capture any dataset bias if an orientation is dominant like in VeDAI). We note that it could give performance improvements to specialize in the most widespread orientations but would hurt generality.
Omitting the regression part, which will be addressed in Section~\ref{reg}, with the IoU computations and the NMS being taken care of, the rest of the implementation is straightforward.

\subsubsection{Oriented RoI pooling}{\label{rroi}}
Augmenting RoI pooling to gain newer abilities has been already tackled by ROIalign \cite{maskrcnn} and several RoI warping pooling have already been proposed \cite{stn,roiwarping,ma2018arbitrary}. 
We favor simplicity and rapidity over precision by using almost the same formulation as the original RoI pooling \cite{fastrcnn}. 
Meaning we use the same harsh quantization to define the pooling cells corners; Then we go through every pixel of the feature map inside the vertical bounding box of the current rotated pooling cell (which identifies to the original non rotated pooling cell when angle is 0) to find the maximum.  

The added complexity lies in the fact that we now have to check for each of these pixels if it belongs to the associated rotated pooling cell. We do that by using simple dot products. 

\begin{figure}[tb]
\begin{center}
\includegraphics[width=0.5\linewidth]{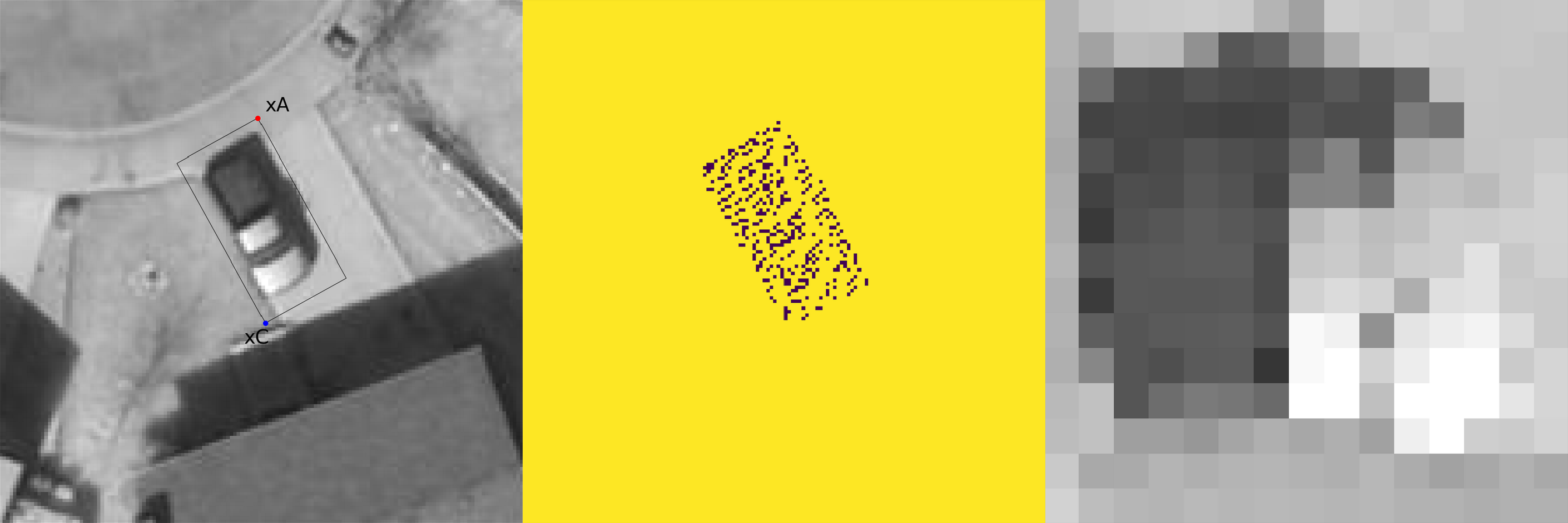}
\caption{From left to right: a Region of Interest and its parametrization on the original image, the locations of the pooled pixels in the original image (argmaxes)(backward), the pooled pixel reordered after Rotated RoI pooling (using the original image for visualization purposes)(forward)}
\label{fig_roi_pooling}
\end{center}
\end{figure}

Obviously due to the coarse quantization of the pixels grid it can happen that a rotated cell is left empty, when it is the case we fill the cell with the nearest neighbor pixel. We provide a visual sanity check of our implementation Figure~\ref{fig_roi_pooling} for a 20x20 region: the quantization is very coarse and we can observe misalignments contrary to other sampler using bilinear interpolation but ours does not need to compute the values of the 4 neighboring pixels.

\subsubsection{Regressing oriented boxes}{\label{reg}}
The vital part of the implementation is the way one regresses the boxes from the anchors (or from the RPN proposals). For instance, if we regress directly the offset ($\Delta_{\theta}$) between our prediction angle $\theta$ and $\theta_{t}$ the target angle in radians, we run into several problems. Indeed by using the simple parameterization described in section~\ref{parametrizing} we have introduced a lot of ambiguity since there are now 4 valid ways of placing A even on a perfect prediction (matching the ground truth exactly) and therefore 4 possible $\Delta_{\theta}$ values associated.  
We have to go through each 4 cases. We follow the same convention of using the subscript t to indicate the ground truth target and no subscript for the original prediction
\begin{itemize}
\item A is near/at $A_{t} \Rightarrow \Delta_{\theta}=0$, $AB \approx A_{t}B_{t}$, $BC \approx B_{t}C_{t}$
\item A is near/at $B_{t} \Rightarrow \Delta_{\theta}=\frac{\pi}{2}$, $AB \approx B_{t}C_{t}$, $BC \approx A_{t}A_{t}$ 
\item A is near/at $C_{t} \Rightarrow \Delta_{\theta}=\pi$, $AB \approx A_{t}B_{t}$, $BC \approx B_{t}C_{t}$ 
\item A is near/at $D_{t}\Rightarrow \Delta_{\theta}=\frac{3\pi}{2}$, $AB \approx B_{t}C_{t}$, $BC \approx A_{t}B_{t}$ 
\end{itemize}
We therefore want gradient in the second and fourth case, otherwise we will go away from the ground truth by regressing $AB$ to $B_{t}C_{t}$, \etc We want no gradient flowing in the other cases as we do not want to force A onto a specific side as it leads to instability and bad generalizations. Of course in addition there is also the problem of the $2\pi$ periodicity.

One can see that using the cost below solves both issues.
\begin{align}
\begin{split}
\Delta_{\theta}&=\theta_{t}-\theta, \\
v&=[\Delta_{\theta}-\pi,\Delta_{\theta}+\pi,\Delta_{\theta}], \\
index&=\argmin(|v|), \\
\Delta_{\theta}&=v[index] \\
\end{split}
\end{align}

Then we can just plug the angle regression term in the regression cost:
\begin{align}
\begin{split}
&RPN=\sum_{a,t} p*\biggl(||\gamma_{1}*(\left(\frac{x_{ct}-x_{ca}}{\frac{W_{a}+H{a}}{2}}\right)-\Delta_{x})||_{\delta}^{2}\\
+&||\gamma_{2}*(\left(\frac{y_{ct}-y_{ca}}{\frac{W_{a}+H{a}}{2}}\right)-\Delta_{y})||_{\delta}^{2}  +||\gamma_{3}*(\log \frac{W_{t}}{W_{a}}-\Delta_{\log w})||_{\delta}^{2}
\\+&||\gamma_{4}*(\log \frac{H_{t}}{H_{a}}-\Delta_{\log H})||_{\delta}^{2}+||\gamma_{5}*((\theta_{t}-\theta_{a})-\Delta_{\theta})||_{\delta}^{2}\biggr)
\end{split}
\end{align}
Where $||.||_{\delta}^{2}$ is the traditional smooth L1 cost with parameter $\delta$. $t$ refers to a target and $a$ to an anchor; the index $c$ marks the center;W is AB and H is BC (see previous sections). And the other unreferenced Greek letters are just hyperparameters (cross validated). 
We integrate this new cost inside the Faster R-CNN framework, we train the RPN and Fast R-CNN parts simultaneously as is done in the literature. %
\section{Experiments}
\subsection{Datasets}
In these experiments, we mainly use the VeDAI dataset \cite{vedai}, which consists of 1200 images of SaltLake City that comes in two different sizes 512x512 and 1024x1024. Every image is available in two versions either colored (RGB) or infrared. The dataset is provided with 10 folds and around 10 classes of vehicles (cars, pickups, vans, boats, etc.). The definition of what is a positive detection is different from the standard Intersection Over Union (IOU) criterion adopted by the Pascal VOC \cite{pascalvoc} or MS COCO \cite{coco}. A detection is considered correct if its center lies within the ellipse centered on the ground truth and lying inside the edges of the target car. %
We are interested in mainly two metrics namely the AP (Average Precision) which is taken on 11 points as in Pascal VOC 2007 and the recall at low FPPI (recall for a given rate of False Positive Per Image).
{All the experiments are performed on SII (Small Infrared Images 512x512)} but the images are first up sampled by a factor 2 using bilinear interpolation and are therefore 1024x1024.

We also use the Munich dataset \cite{munich} to verify that our framework is adapted to various aerial imagery settings. This dataset contains 20 large colored images. 10 for training and 10 for testing. The images have to be rasterized for efficient training. We thus cut the images in a grid of 512x512 patches that we up-sample to 1024x1024 using bilinear interpolation. Doing so, we have missed some of the vehicles. We sample additional 512x512 windows so that each ground truth can be found in at least one image. For test we perform sliding window with a step of 50 pixels, as in \cite{Tang2017}, and we resize each window.
By resizing we augment the computational cost but we avoid having to use higher resolution feature maps.

Finally, we use the Google Earth dataset provided by Heitz and Koller \cite{google_earth} augmented by the annotations of Joao and Vedaldi \cite{warped}. There are 20 colored images taken over the city of Bruxelles. We train on 10 of them and test on 5 other as in \cite{warped}. As for Munich we use Pascal VOC IOU and VeDAI metric to ascertain the quality of the detection. We also have to do some up sampling on GoogleEarth otherwise neither Faster nor our framework converges.

\subsection{Training details}
We use stochastic gradient descent with a learning rate of 1e-3 and momentum for our framework; for Faster R-CNN it is actually too big of a learning rate and does not converge on several folds therefore we used 0.0001 instead. We validated hyperparameters for Faster R-CNN and our framework on fold01 validation set (VeDAI). Our framework has a grid with 36 anchors in each cell 9 angles two aspect ratios 0.5 and 1. and two sizes. We also experimented with anchors obtained by applying kmeans++ \cite{kmeans} to angle normalized and centered anchors of the ground truth with $1-IOU$ distance like in Yolo v2~\cite{yolo9000} but found that it only gave a small boost w.r.t. a reasonable choice of base anchors. This is not entirely surprising considering the lack of variability in the aspect ratio in aerial imagery evoked in the introduction. For Faster R-CNN we also used as many as 36 anchors\footnote{We use a dense sampling of aspect ratio and sizes for Faster R-CNN anchors to get 36 anchors (6 and 6 respectively)} to verify that we were not simply outperforming Faster R-CNN because we had more anchors per pixel. We use for both framework the same vgg16 \cite{vgg} backbone pretrained on IMAGENET.

\subsection{Experimental validation of the Faster RER-CNN}

All subsequent experiments in this subsection are conducted on {fold1 test of VeDAI}, with the objective of validating that everything works as expected and producing a comprehensive analysis of our algorithm. During comparisons with state-of-the-art methods (next section) we will use the entire dataset.

\subsubsection{Comparison of the Region Proposal Networks with and without rotated anchors}

\begin{figure}[tb]
\begin{center}
\includegraphics[width=0.9\linewidth]{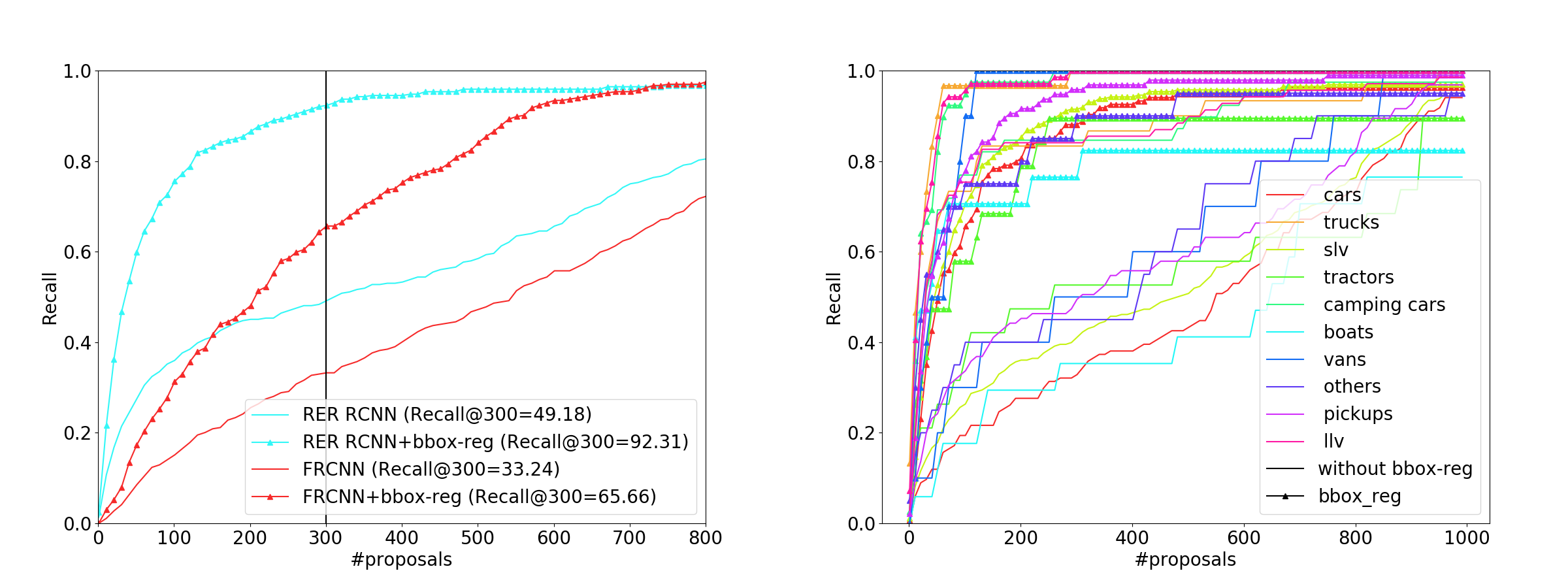}
\caption{Left: comparison of the mean Recall of the RPN across classes between both frameworks, Right: Per class recall of the RPN for our framework}
\label{fig_rpn_vs}
\end{center}
\end{figure}

In these experiments, we want to demonstrate how important it is to have rotated anchors in the region proposal step. In order to do this, we measure, on the test set, the mean recall across classes on the VeDAI dataset w.r.t. the VeDAI metric (see Section~\ref{metrics}) as a function of the number of proposals generated. To do this we extract from the trained network the top N proposals from the RPN (after NMS) and check how many objects are hit. We do this \textit{before regression} therefore the only possible locations of the proposals lay on the original grid \textit{and after regression} when proposals are allowed to move and warp. Our framework, of course, it was built for that purpose, offers more flexibility as the boxes can rotate. It is clear when looking at Figure \ref{fig_rpn_vs} the advantages that the rotated anchors bring. What is surprising is that even for few proposals (400) the gaps between the different RPNs are well marked: from 65.66 to 92.31 (a 41\% increase!) after regression. Even more surprisingly, even before regression the recall is higher (by 48\% !) meaning that since the metric is solely based on the position of the centers of the proposals and that all centers lie on the same grid for both frameworks it must be that training has been made easier/more general by our formulation and that better anchors are chosen. On the right-hand side of Figure~\ref{fig_rpn_vs} we can see the details per class for our framework.

\subsubsection{Angle regression in the classification head}
\begin{figure}[tb]
\begin{minipage}[b]{.4\textwidth}
\centering
\includegraphics[width=1\textwidth, height=0.6\textwidth]{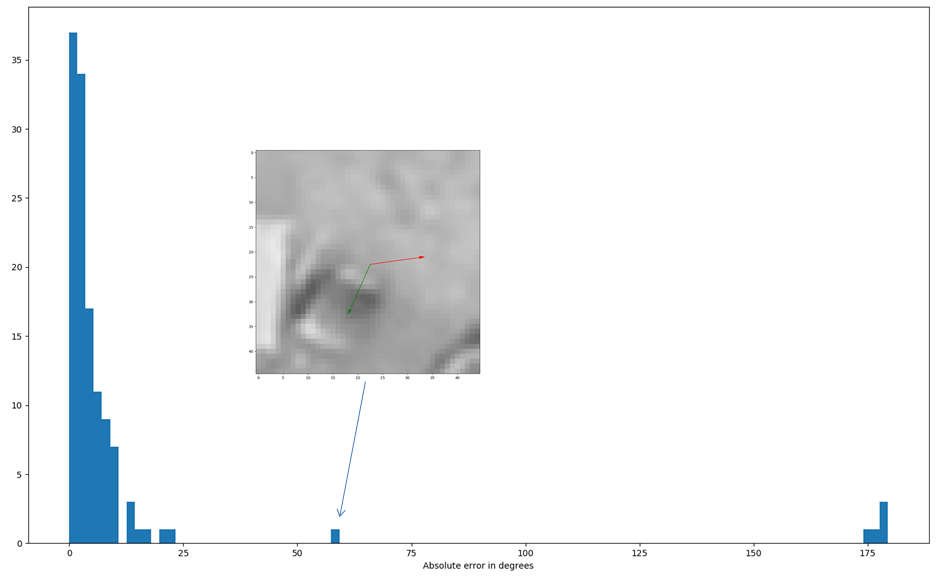}
\caption{Histogram of the error of the angle prediction \label{hist}}
\end{minipage}
\hfill
\begin{minipage}[b]{.3\textwidth}
\centering
\includegraphics[width=1\textwidth, height=0.8\textwidth]{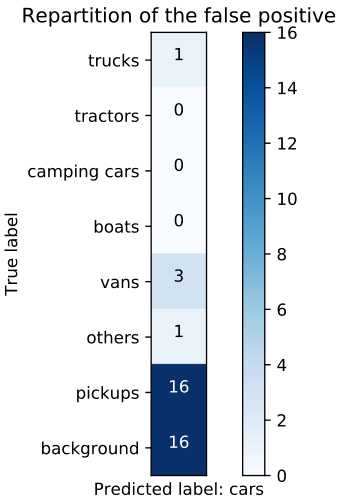}
\caption{Confusion vector of the "car" class detector\label{conf}}
\end{minipage}
\end{figure}

Now that we have demonstrated  our RPN is working, we want to check that the regression of angle is functioning properly in both the RPN, which seems to be the case according to the previous section, and the classification-regression branch on top of it, which we have yet to prove. We display a histogram of the repartition of the absolute error between all final detections (after rescoring and regression of the proposals by the Fast R-CNN branch) labeled as true positives (wrt the VeDAI metric) and their corresponding ground truth on Figure \ref{hist}.

Only one patch is detected with more than 25 degrees of error and most of the other errors are negligible ($<10$ degrees). We note that Liu and Mattyus \cite{Liu15:fast} also got good results of angle estimation on Munich as the orientation of a rectangle is easy enough to find our framework is cleaner as it does detection and angle regression simultaneously. The only big error ($>40$ degrees) comes from a more than questionable ground truth (visible on Figure \ref{hist} too). We draw the reader's attention on the fact that an error of $\pi$ is not an error in our framework.

\subsubsection{Rotation equivariance of the proposals}

\begin{figure}[tb]
\begin{center}
\includegraphics[width=0.7\textwidth]{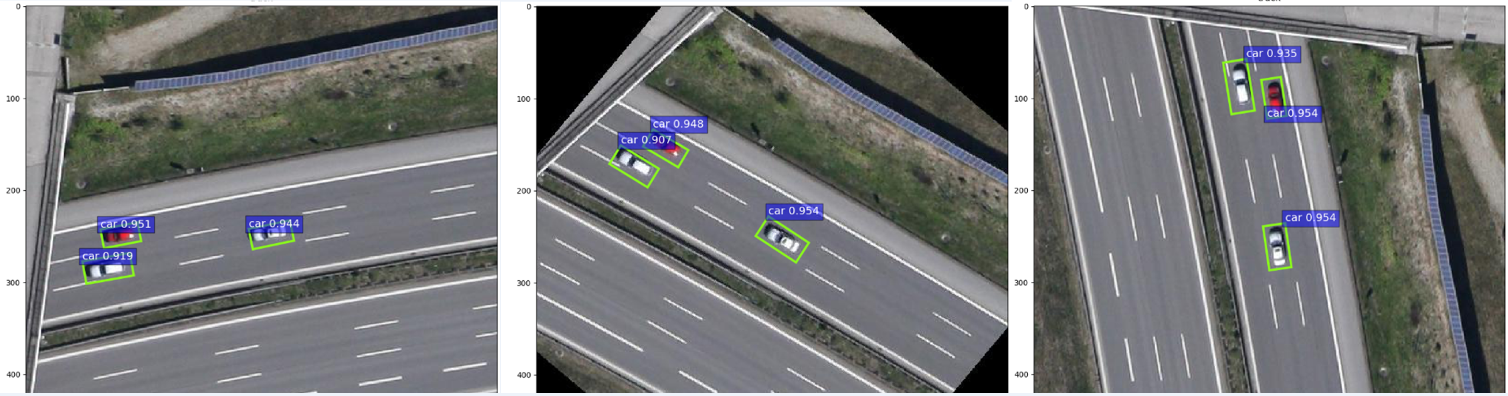}
\caption{Detections given by  our Faster RER-CNN w.r.t. rotation}
\label{rot_munich}
\end{center}
\end{figure}

We do additional experiments on Munich test set by checking the performances of the detector on series of rotated images and see how the recall and precision evolve. It turns out there is little change in the performance when the images are rotated. A subset of the detections is presented Figure \ref{rot_munich}. The experiments show that our detector is, to a certain extent, provided that the RPN outputs correct proposals, rotation-equivariant: when one rotates the image the detections undergo the same transformation. One downside is that hard backgrounds remain hard backgrounds whatever the orientation.

\subsubsection{Confusion analysis of the 'car' detector on VeDAI}
\begin{figure}[tb]
\begin{minipage}[b]{.5\textwidth}
\centering
\includegraphics[width=1\textwidth]{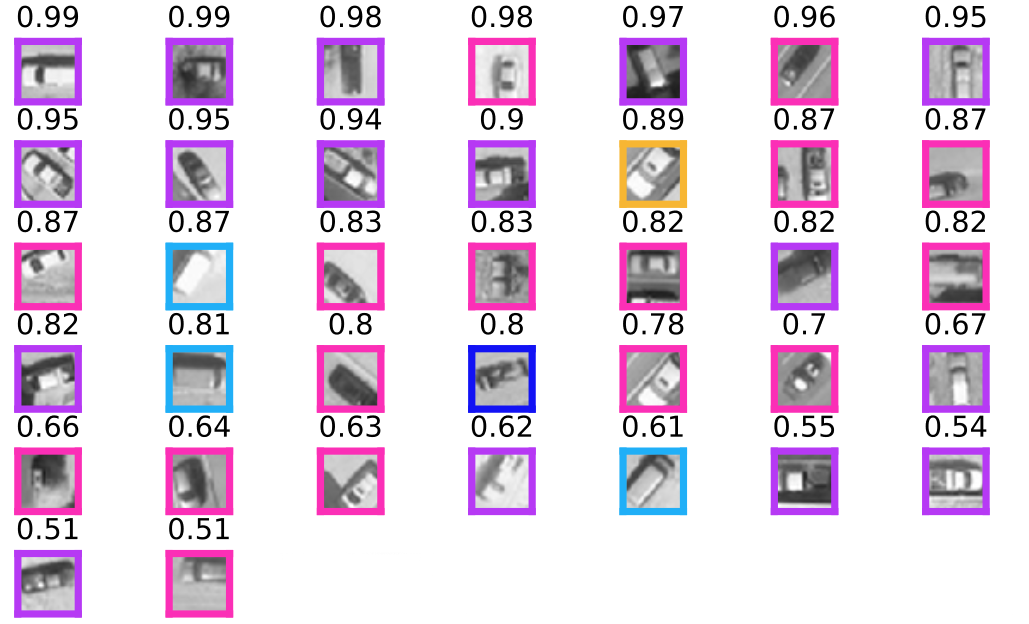}
\caption{False positives for the car detector with scores$>0.5$ The color code used is vans: turquoise, trucks: dark blue, pickups: purple, others: yellow, backgrounds: pink\label{patches}}
\end{minipage}
\hfill
\begin{minipage}[b]{.4\textwidth}
\centering
\includegraphics[width=0.8\textwidth]{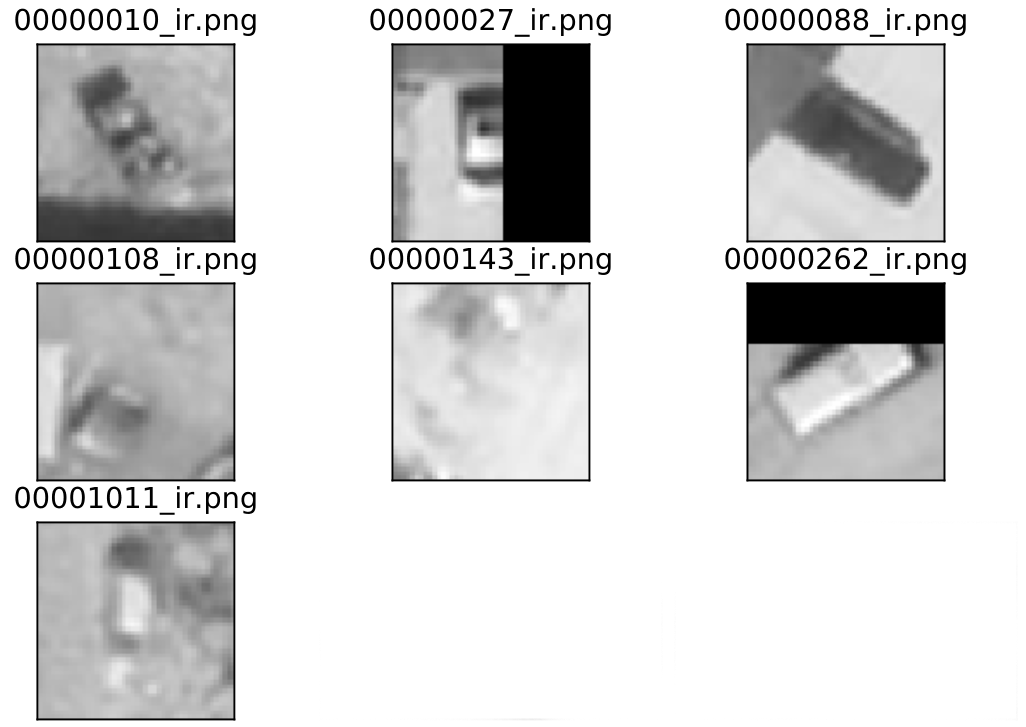}
\caption{The only 7 ground truths {labeled as cars} never found by our car detector (out of 121 in fold01), numbers 4 and 5 do not resemble cars and must be annotator mistakes and numbers 2,6 and 7 are heavily occluded \label{misses}}
\end{minipage}
\end{figure}

{\label{sec:confusion}}
We now focus on {the car class detector} extracted from our framework trained on VeDAI and display the results of a thorough miss and false positive analysis. The results of this study can be seen first on Figure~\ref{conf}, where we show the repartition of all the false positive whose score are superior to 0.5 (it is hence a column vector part of a confusion matrix), Figure \ref{patches} shows the associated false positive patches (non-scaled and non-rotated). It is obvious that even most of the background mistakes (in pink) are just localization errors on other similar classes. With this kind of resolution, it is often hard even for a human to distinguish between say a pick-up and a car. We also display {every} car ground truth missed by our detector no matter the threshold applied in Figure~\ref{misses}.

It is now time to test it more thoroughly on several benchmarks and to compare it against different baselines and the state-of-the-art.

\subsection{General detection results}

\subsubsection{A note on the evaluation protocol}{\label{metrics}}
As stated in the introduction comparing detection methods when their results and ground truth are not the same is not self-evident. The problem of intercepting a rectangle with another is obviously more difficult if you allow both rectangles to rotate. One solution would be to use PASCAL VOC vertical boxes as ground truth for both methods but it would be unfair towards our framework, which was trained on rotated boxes only (it would also be unfair towards Faster R-CNN to use the rotated boxes as ground truth even if they are evidently closer to the reality). Therefore in addition to VOC criteria for each framework on their respective ground truth we provide an absolute measurement, which is the VeDAI metric (defined in \cite{vedai}), which only takes into account the position of the center of the predictions and thus treats each solution equitably. 

\subsubsection{VeDAI results}
\begin{figure}[tb]
\begin{center}
\includegraphics[width=0.8\linewidth]{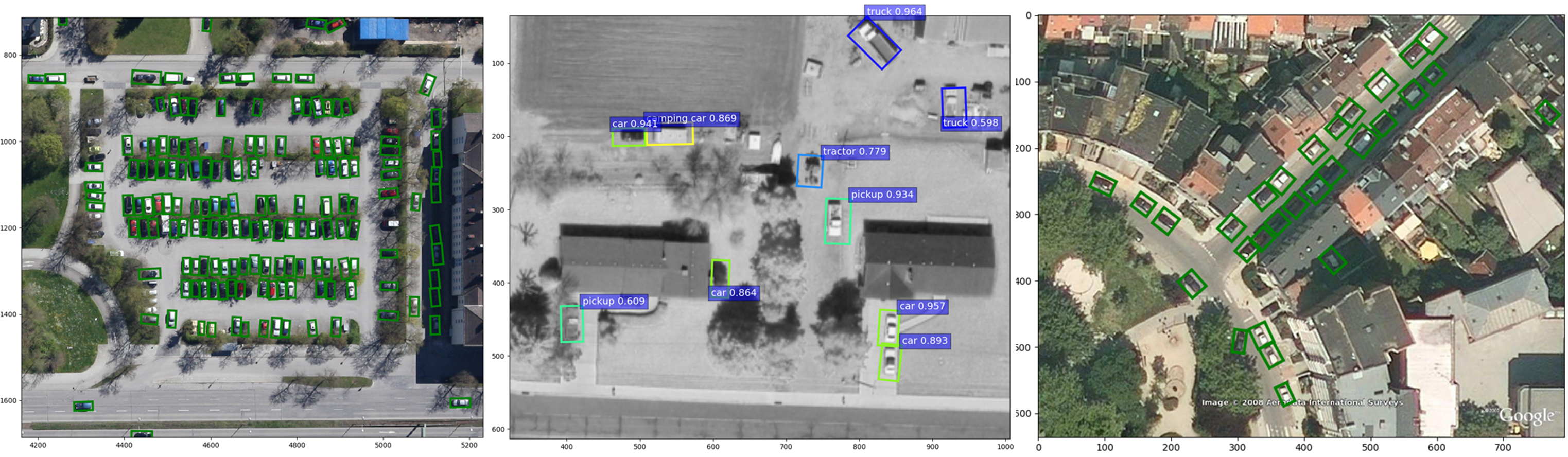}
\caption{Example detections on all 3 datasets from left to right: Munich3k, VeDAI SII, GoogleEarth}
\label{im_results}
\end{center}
\end{figure}

{\label{vedai_results}}
We have detection results on all folds and all classes of VeDAI. We first compared with related works on the car class (the only one present in \cite{me}). Then we move to a detailed analysis per class for both frameworks. The only metric presented is the one described in Section \ref{metrics}.
Some samples detection from VeDAI zoomed in can be seen in the center of Figure \ref{im_results}.

\begin{table}[tb]
\centering
\resizebox{0.4\linewidth}{!}{%
\begin{tabular}{|c|c|c|}\hline
method&AP&Recall@0.01FPPI\\ 
\hline
DPM \cite{vedai} &$60.5 \pm 4.2$&$13.4 \pm 6.8$\\
SVM+HOG31 \cite{vedai}&$55.4 \pm 2.6$&$7.8 \pm 5.5$\\
SVM+LBP \cite{vedai}&$51.7 \pm 5.2$&$5.5 \pm 2.2$\\
SVM+LTP \cite{vedai}&$60.4 \pm 4.0 $&$9.3 \pm 3.7$\\
SVM+HOG31+LBP \cite{vedai}&$61.3 \pm 3.9 $&$8.3 \pm 5.2$\\
SVM Fusion AED (HOG) \cite{sebastien2}&$69.6 \pm 3.4 $&$20.4 \pm 6.2$\\
FCN \cite{me} &77.80 $\pm$ 3.3& 31.04 $\pm$ 11\\
Faster R-CNN (our implementation)  &77.69 $\pm$ 3.4& 20.85 $\pm$ 14 \\
Ours  &\textbf{80.2} $\pm$ \textbf{3.3}&\textbf{33.2} $\pm$ \textbf{14} \\
\hline
\end{tabular}}
\caption{Comparisons with related works on VeDAI\label{table:results}}
\end{table}

One can see in Table~\ref{table:results} that our work outperforms all others including the recent state of the art of \cite{me} that uses hard example mining. The associated precision-recall curves for the car class are displayed Figure \ref{pr}.

Table \ref{table:10foldsvedai} shows that indeed a big enhancement is observed on all the vehicles (except boats, camping cars and planes where the two reported numbers are about equal) proving our point that our framework is beneficial for all  vehicles. There is a huge variance on the plane class as there is very few of them. The variance is not negligible in all the other classes due to the very unequal difficulty of the different folds.  For more explanation on confusion between classes, see section \ref{sec:confusion}.

\begin{table}[tb]
\begin{minipage}{0.4\linewidth}
\centering
\resizebox{1\linewidth}{!}{%
\begin{tabular}{|c|c|c|c|}\hline
class & F. R-CNN & FCN\cite{me}& Ours \\ 
\hline
cars & 77.69 $\pm$ 3.4 & 77.80 $\pm$ 3.3 & \textbf{80.2 $\pm$ 3.3} \\
\hline
camp. cars & 72.72 $\pm$ 5.6 & --- & \textbf{72.8 $\pm$ 4} \\
\hline
pickups & 74.84 $\pm$ 2.77 & --- & \textbf{77.0 $\pm$ 4.2} \\
\hline
tractors & 54.43 $\pm$ 7.00 & --- & \textbf{67.8 $\pm$ 12.1} \\
\hline
trucks & 66.73 $\pm$ 9.97 & --- & \textbf{72.3 $\pm$ 7.6} \\
\hline
vans & 69.93 $\pm$ 9.13 & --- & \textbf{74.1 $\pm$ 11.0} \\
\hline
boats & \textbf{66.22 $\pm$ 7.99} & --- & 65.3 $\pm$ 8.3 \\
\hline
others & 43.39 $\pm$ 11.45 & --- & \textbf{51.0 $\pm$ 8.8} \\
\hline
planes & \textbf{77.85 $\pm$ 34.45} & --- & 77.4 $\pm$ 32.7 \\
\hline
mean & 67.09 $\pm$ 13.6 & --- & \textbf{70.88 $\pm$ 13.3} \\
\hline
\end{tabular}}
\caption{Comparisons baselines VS our framework on VeDAI (10 folds)\label{table:10foldsvedai}}
\end{minipage}\hfill 
\begin{minipage}{0.4\linewidth}
\centering
\includegraphics[width=0.8\linewidth,height=0.6\linewidth]{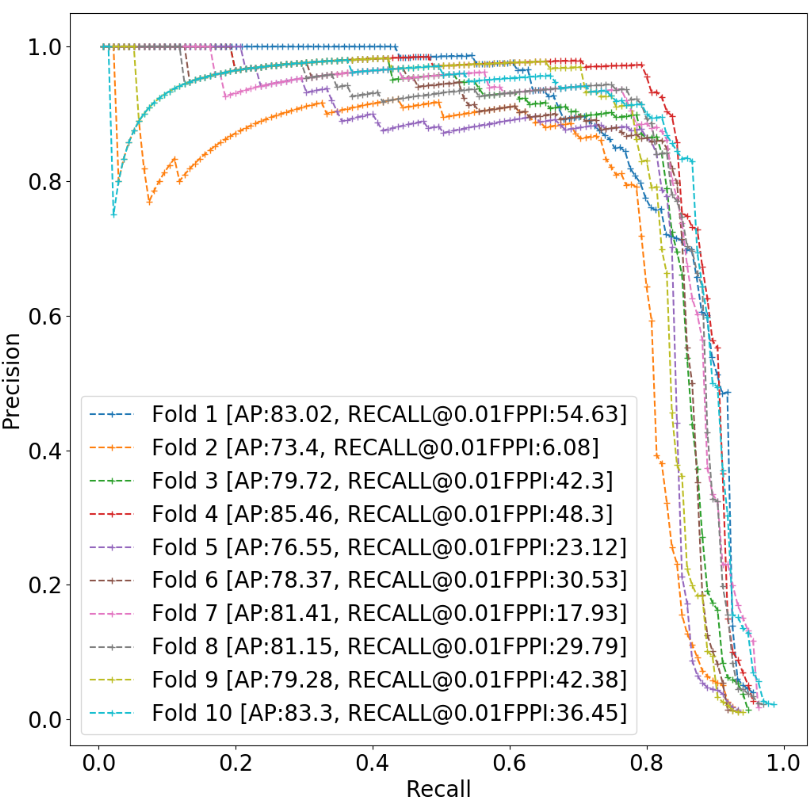}
\captionof{figure}{PR curbs for the car class on VeDAI 10 folds \label{pr}}
\end{minipage}
\end{table}

\subsubsection{Munich 3K results}
On Munich \cite{munich} several tricks need to be applied to match the state of the arts such as using better networks, hard example mining or increasing the resolution of the feature map. We did not go that far as it was not the purpose of this article. %
Results presented in Table \ref{table:results_munich} speak for themselves.

\begin{table}[tb]{%
\begin{center}
\begin{tabular}{|c|c|c|}\hline
method&AP VOC@0.3 & AP VEDAI\\ 
\hline
Faster R-CNN \cite{fasterrcnn}& 85.59 & 85.68 \\
Ours  &\textbf{87.14} & \textbf{87.32}\\
\hline

\end{tabular}
\end{center}}

\caption{Comparison Faster RCNN vs framework on Munich3k}\label{table:results_munich}
\end{table}

We observe that contrary to VeDAI, PASCALVOC@0.3 and VeDAI are in accordance because in both datasets most errors are not due to precision of localization of the boxes but to hard backgrounds and hard to find vehicles

\footnote{It is worth going through the supplementary to find detection examples of all datasets in color and magnified.}.

\subsubsection{GoogleEarth results}
\begin{figure}[tb]
\begin{center}
\includegraphics[width=0.5\linewidth]{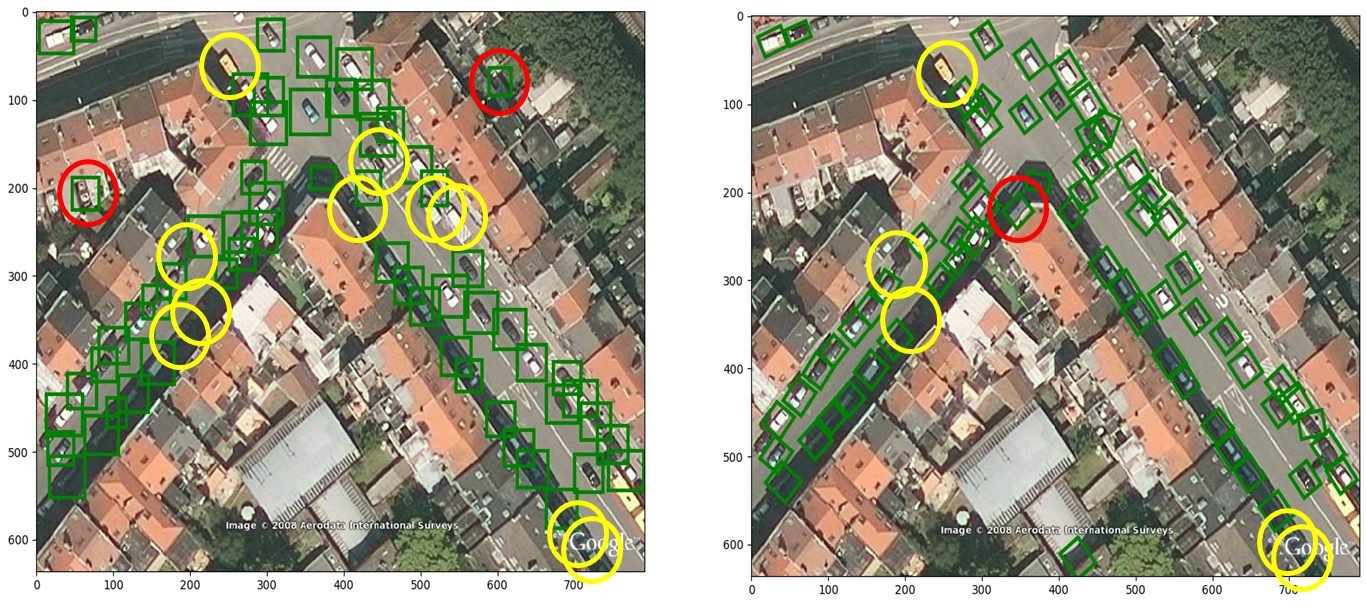}
\caption{Vehicles missed by the detector are indicated with yellow circles and false positives in red. Left: Faster R-CNN. Right: Faster RER-CNN. This figure shows that although the problem remains difficult our framework has higher precision and recall and gives a lot more information about the vehicle's templates.}
\label{google_images}
\end{center}
\end{figure}

We tested on this very small dataset only to show the generality of our solution for aerial imagery problems. However the number of images and vehicles is too small thus we have probably overfitted the data with either Faster R-CNN or Faster RER-CNN. Nonetheless again even if the disparities are small our framework is still leading in performances as can be seen on Table~\ref{table:results_google}. Figure~\ref{google_images} shows one of the 5 test images and the errors w.r.t. a 0.5 detection threshold of both frameworks side by side. 

\begin{singlespace}
\begin{table}[tb]{%
\begin{center}
\begin{tabular}{|c|c|c|}\hline
method&AP VOC@0.5 & AP VEDAI\\ 
\hline
Faster R-CNN \cite{fasterrcnn}& 84.81 & 87.37 \\
Ours  &\textbf{88.39} & \textbf{88.53}\\
\hline
\end{tabular}
\end{center}}
\caption{Comparison Faster RCNN vs framework on GoogleEarth}\label{table:results_google}
\end{table}
\end{singlespace}
\subsubsection{Running Times}
We also add in table \ref{table:speed} speed comparisons of our framework w.r.t. the baseline. The bottleneck of our implementation being the IoU computations and NMS done with the CPUs (nms is on GPU for Faster) we expect to see a slowdown. It takes approximately twice as long to train and test, which is reasonable given that we get the angle and template.
\begin{singlespace}
\begin{table}[tb]{%
\begin{center}
\begin{tabular}{|ccc|}\hline
method& training step (s) & inference(s)\\ 
\hline
Faster R-CNN \cite{fasterrcnn}& 0.362 & 0.158 \\
Ours  & 0.602 & 0.365\\
\hline
\end{tabular}
\end{center}}
\caption{Running times comparisons for one image 1024x1024 on an Intel i7-7700K CPU@8*4.2GHz with a GTX1080 (averaged on 100 runs)}\label{table:speed}
\end{table}
\end{singlespace}

\section{Conclusions}
We have presented a simple but strong alternative for the detection of small vehicles in aerial imagery built from Faster R-CNN architecture and based on rotation invariance of the classification. We saw how using rotated boxes in the RPN increases recall while maintaining precision allowing us to beat other alternatives on several benchmarks. It is our hope that this framework will serve as the new reference baseline on all the standard benchmarks from aerial imagery. Furthermore by simultaneously predicting the box and the angle we get {for free} a completely tight box matching the vehicle exact dimensions. It could therefore be used in a third stage to compare with a database of known vehicle dimensions (the ground to camera distance  is known and the views are vertical) to get a more precise discrimination between vehicles, which is probably the hardest part. As a side note we also implemented a rotated version of RoI Align \cite{maskrcnn} -- it is completely straightforward to integrate inside our framework -- and we are currently exploring it as a way to further increase the accuracy of the method at the expense of speed.

\bibliographystyle{ieee}
\bibliography{references}

\end{document}